\documentclass[conference]{IEEEtran}
\ifCLASSINFOpdf
  \usepackage[pdftex]{graphicx}
\else
\fi
\usepackage{amsmath}
\usepackage{amsfonts}
\usepackage{mathtools,xparse}
\usepackage[section]{placeins}

\DeclarePairedDelimiter{\norm}{\lVert}{\rVert}
\NewDocumentCommand{\normL}{ s O{} m }{%
  \IfBooleanTF{#1}{\norm*{#3}}{\norm[#2]{#3}}_{L_2(\Omega)}%
}
\begin{document}
%
\title{Deep Learning: Generalization Requires Deep Compositional Feature Space Design}

\author{\IEEEauthorblockN{Mrinal Haloi}
\IEEEauthorblockA{Indian Institute of Technology, Guwahati\\
h.mrinal@iitg.ernet.in}
}


%


\maketitle

\begin{abstract}
Generalization error defines the discriminability and the representation power of a deep model. In this work, we claim that feature space design using deep compositional function plays a significant role in generalization along with explicit and implicit regularizations. Our claims are being established with several image classification experiments. We show that the information loss due to convolution and max pooling can be marginalized with the compositional design, improving generalization performance. Also, we will show that learning rate decay acts as an implicit regularizer in deep model training.\\ 

Keywords: Generalization, Deep Compositional Design, Convolutional Network, Deep Learning  
\end{abstract}


%
\IEEEpeerreviewmaketitle

\section{Introduction}
Deep learning massive success in almost every fields represents its ability to solve complex problems. The trade-off between model complexity and accuracy is an important area of deep learning research. Very complex model with millions of parameters \cite{inceptionv4, resnet} proved to the state of the art solution for many vision and natural language problems. A common way to measure the performance or generalizability of a deep learning model is to test it on a well discriminative validation/test set representing the variation of samples of the corresponding problem. Learning very complex model is a matter of the requirements of high computing power and huge dataset. So it's important to understand the optimal complexity requirement for a problem to reduce the burden of computing power. In a recent work by Zhang et al. \cite{gb_general}, it has been proved that a simple 2 layers neural network with $2n+d$ parameters can represent any function for $n$ samples in $d$ dimensions. It is interesting to see that a simple multilayer perceptron with ReLU activation can fit a dataset of random labels with zero training accuracy, but poor generalizability. The problem with mere data memorizing is to be blamed for poor performance on a test set. Preventing the network in memorizing data samples in inefficient random high dimensional space is important model design paradigm. But at the same time, it's acceptable for a model to memorize the data in an efficient hyperspace representing original data distribution. Learning the original data distribution solves the poor validation set performance of a model. Designing an optimal network with a minimum number of parameters will reduce computation costs and improve performance. \\
  How do we design model that best fit the original data distribution? In this work, we will present the importance of deep compositional feature space design with an optimal number of parameters. We will prove the rate of feature space size reduction matters irrespective of network parameters. Also, we will define an optimal strategy to relate a number of parameters requirements for a particular feature space representation.
\subsection{Contribution}
\textbf{Feature space representability:} There exists an optimal number of nonlinear transformations to represent a particular size features space without losing discriminative information. Convolution is a linear operation projecting data from one space to another, using a nonlinear activation final output of convolution transformation becomes non-linear. While transforming features from one space to another space there's loss of information. The loss of information can be understood using singular value decomposition. We have trained multiples network with the same number of parameters but a different rate of feature space reduction on the CIFAR10 dataset to prove our points. Following observations have strong impact on the learning and the generalization performance:
\begin{itemize}
  \item{The rate of reductions of the feature space size with respect to a number of convolution operations.}
   \item{onvolution vs max pooling for feature space reduction.}
 \end{itemize}
\textbf{An optimal number of model parameters:} There exists an optimal number of model parameters for a model to achieve high generalization accuracy. The number of optimal parameters depend on the compositional design and the feature space reduction rate.\\
\textbf{Implicit Regularizer:} Learning rate decay policy acts as an implicit regularizer boosting the model performance and faster learning.\\
Note: here with the term feature space size, we talk about width and height of a feature map, not the number of feature maps.

\subsection{Related Work}
Zhang et al.(2017)\cite{gb_general} studied the representational power of a network with respect to the training sample size; shows that a deep model can memorize any dataset with random labels, but it doesn't imply generalization. Their finding also stated that explicit regularization alone can't prevent poor generalization performance.
Barlett (1998)\cite{barlett} showed that VC dimension is not relevant to measure generalization performance for neural networks, rather the $L1$ norm of the network weights is more prevalent measure. They defined a fat-shattering dimension for error estimation that depends on parameter magnitude. Maass (1995)\cite{maass} has established the VC dimension bounds for the neural network with various activation functions for generalization analysis. Krogh et al. (1992)\cite{krogh} showed that weight decay suppresses any irrelevant weights vector component also reduce noises, hence lowering generalization errors. Hardt et al. (2016)\cite{hardt} introduced a generalization error upper bound for a model trained with stochastic gradient descent for convex and non-convex optimization problems. Neyshabur et al. (2015)\cite{neyshabur} shows that apart from adding $L2$ weight decay or implicit regularization, increasing the network size improves generalization performance for a model learned with stochastic gradient descent. They also asserted that with very high number of hidden units ($> sample size$) a weight decay regularized network is considered as a convex neural net for optimization. 
On the effectiveness of deep networks vs shallow networks Mhaskar et al. (2016)\cite{mhaskar}, showed that VC dimension and fat-shattering dimension are smaller for deep networks than shallow networks. They argued the benefits of compositional function design for scalability and shift invariance in image and text data. All these above works didn't discuss the prominent effect of the loss of information while data is projected from one space to another. We will show here that the information loss effect the generalization error for deep or shallow networks.   

\section{Feature space design in deep neural networks}
\label{fp_design}
\subsection{Information loss}
The standard convolution operation used in the deep network is linear. When feature space is projected from one space to another using convolution/pooling operation there's always a loss of information. The information loss depends on the projected space dimension and capacity. Loss of information can be understood from singular value decomposition (SVD). If $F_{i}$ denotes the input for $i^{th}$ convolutional layer
\begin{equation}
\label{eq:svd}
F_{i}^{j}F_{i}^{jT} = WEW^{T}
\end{equation}
\begin{equation}
F_{i, proj}^{j} = W_{d}^{T}F_{i}^{j}
\end{equation}
$F_{i}^{j}$ is the $j^{th}$ input map, the equation~\eqref{eq:svd} refer to a special case where the input matrix $F_{i}^{j}F_{i}^{jT}$ for SVD is Hermitian and positive definite. When data is projected from $n$ dimension to low-rank approximation $d < n$ dimension, there's loss of information. 2-D convolution for a single feature map with a single kernel can be interpreted as projecting data from one space to another. Significant information is lost when convolution stride $>$ 1. Another way to calculate the retrived information after convolution is using the mutual information between two signals. For two independent signals X and Y, mutual information I(X, Y) = 0 and maximum if X $\approx$ Y (fully correlated). For X and Y we define information loss as follows:
\begin{equation}
\label{eq:inf_loss}
information_loss \propto \frac{1}{I(Y; X)}\\
\end{equation} 

 Mutual information between the original data and projected data (convolution/pooling) is given as follows
\begin{equation}
\label{eq:mutual_inf}
I(F_{i, proj}^{j};F_{i}^{j}) = H(F_{i}^{j}) - H(F_{i}^{j}|F_{i, proj}^{j})
\end{equation}
For $N-D$ normal random vectors X $\sim N(\mu_{X}, C_{X})$ and Y $\sim N(\mu_{Y}, C_{Y})$ mutual information can be calculated as follows:
\begin{equation}
\label{eq:inf_loss_deriv}
\begin{aligned}
I(X;Y) &= H(X) - H(X|Y) \\
I(X;Y) &= \frac{1}{2}log(\pi e)^{N}log(\frac{\lvert C_{X} \rvert \lvert C_{Y} \rvert}{\lvert C \rvert}) \\
C &= \begin{bmatrix}
C_{X} & C_{XY}\\
C_{YX} & C_{Y}
\end{bmatrix}
\end{aligned}
\end{equation}

Convolution of a $2-D$ image with $n\times n$ kernel is equivalent to the same with $1\times n$ and $n \times 1$ kernels; 2 1-D convolution.

Where $C_{*}$ are covariance matrices for respective variables. Since we are using batch normalization also input whitening, it's safe to assume input/output of convolution as normal. Apart from that convolutional kernel also initialized as normal variables. 
If the convolution input X is sampled from a normal distribution X $\sim N(\mu_{X}, C_{X})$ and the kernel K $\sim N(\mu_{K}, C_{K})$, then the ouput will also be a normal with distribution Y $\sim N(\mu_{X}+\mu_{K}, C_{X}+C_{K})$. Using eq~\ref{eq:inf_loss} and eq~\ref{eq:inf_loss_deriv} we can calculate information loss for convolution/pooling. 

\subsection{Compositional Design of Convolutional Layer}
A single convolution layer followed by max pooling results in high information loss. Increasing the number of convolution kernel for a layer doesn't necessarily solve the problem. Before reducing the feature space size it's important to project the feature into high dimensional hyperspaces using multiple convolution operations. It's important to use non-linearity and batch normalization \cite{bn} for each convolution operation to achieve highly uncorrelated hyperspace projection. The stacking design best resembles the representation power of the compositional function. Also, the VC dimension of compositional design is smaller than that of shallow design \cite{mhaskar}. With the composition of multiples convolution operation, receptive field grows in polynomial order. Compositional design inspired from the visual cortex increasing receptive fields for higher visual areas. It has been established \cite{cortex} that visual cortex receptive fields are larger for a simple scene and smaller for a complex scene. Compositional design has smaller and larger receptive fields those capture information related to simple and complex objects leading to decrease in information loss.

\begin{equation}
\label{eq:shallow}
\begin{aligned}
f_{shallow}(input) &= conv->bn->relu(input)\\
receptiveField &= filtersize
\end{aligned}
\end{equation}
\begin{equation}
\label{eq:composition}
\begin{aligned}
H &= conv->bn->relu\\
f_{composition}(input) &= H(H(H(H(input))))\\
receptiveField &= 1 + 4(filtersize - 1)
\end{aligned}
\end{equation}
The above equations~(\eqref{eq:shallow}, \eqref{eq:composition}) only valid for convolution stride 1. 

\subsection{Convolution vs Max Pooling}
\begin{figure}
  \centering
      \includegraphics[width=3.1in,height=2.2in]{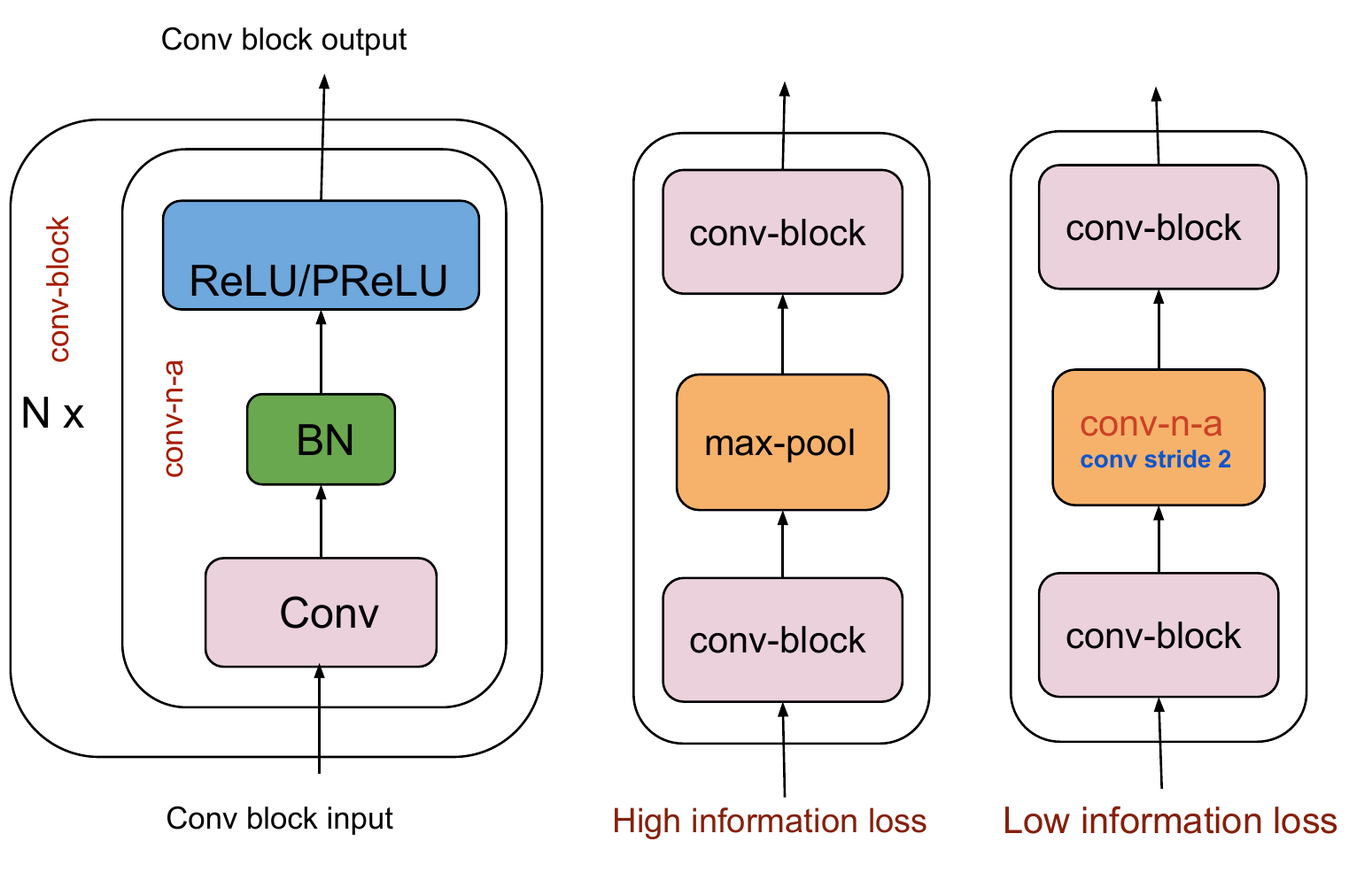}
\caption{Left: A convolutional block design with compositional convolutional operations (conv-n-a means convolution followed by normalization and non-linearity). Middle: design with max-pool for downsampling. Right: convolution for downsampling}
\label{fig:conv_pool}
\end{figure}

\textbf{Claim 1.1} Strided convolution replaces max/avg pooling with better generalization performance.\\
Feature space reduction using max pooling is a very crude projection into another hyper space. Max pooling operation leads to lossy non-linear transformation. The translational invariance which is one of the advantages of max pooling operation can be easily well represented using affine transformation, achieved with strided convolution. Information loss in strided convolution is lower than hard non-linear max pooling. Fig~\ref{fig:conv_pool} illustrates model design with strided convolution and max pooling.\\
 Table \ref{tab:ac_max_conv} proves our claims. 

\subsection{Rate of Reduction}
\textbf{Claim 1.2} Minimum one convolutional operation needed before reducing the feature space size of the model.\\
To minimize loss of information, projecting the data into mutliples non-linear hyper space is required for improved generalization. Minimum one convolution for the first layer of the model without stride and maximum 4 convolution operations without residual connection is preferable for intermediate layers; with residual connection, $ > 4$ convolutional operations can be added. It is also noteworthy that residual connection only facilitate the training of deeper model.

\section{Generalization requires feature space analysis}
\subsection{VC dimension and Fat Shattering dimension}
A function $f: [a, b]$ is considered as Lipschitz function if it satisfies the following condition for a smallest constant $c$:
\begin{equation}
\label{eq:lipschitz}
|f(x)- f(x^{'})| \leq c|x-x^{'}| \hskip 2em \forall x,x^{'} \in [a, b]
\end{equation}
For deep networks non-linearity like ReLU is lipschitz function for $x \in [0, \infty]$ and sigmoid for $x \in \mathbb{R}^{n}$

\emph{VC} dimension for a neural network class H with $l$ layers, inputs $X \subset \mathbb{R}^{n} $ and ReLU activation function is given as follows \cite{maass}
\begin{equation}
\label{eq:vc}
\begin{aligned}
lim_{x \rightarrow \infty}relu(x)& \neq lim_{x \rightarrow -\infty}relu(x) \\
x & \in \mathbb{R} \hskip 2em relu'(x) \neq 0 \\
VCdim(Net) &= O(w\emph{l}logw + wl^{2})
\end{aligned}
\end{equation}
Significance of VC dimension analysis for deep convolutional neural network training is marginal. The theretical $O(n^{2})$ complexty for $n$ number of total parameters of convolutinal model is rather very high upper bound for consideration in generalization analysis. It has been established that implicit and explicit regularization improves generalization. 
\emph{Fat Shattering} dimension of a neural network class G with $l$ layers and inputs $X \subset \mathbb{R}^{n} $ is given as follows \cite{barlett1} for some constant $c$ and $\lambda > 0$; number of points $\lambda$-shattered by G
\begin{equation}
\label{eq:fat}
\begin{aligned}
fat_{G}(\lambda) &= O(\frac{B^{2}(cA)^{l(l+1)}}{\lambda^{2(l-1)}}) \\
X &= {x \in \mathbb{R}^{n}; \lVert x \rVert_{\infty} \leq B} \\
\lVert w \rVert_{1} & \leq A
\end{aligned}
\end{equation} 
Fat-shattering dimension is better bound than VC dimension for learning algorithms, as it suggets that minimizing the values of networks parameters is important for better generalization. The values of model parameters can be minimized using $L1/L2$ regularizer; achieved adding a extra penality term to the cost function using respective norms. 
\begin{equation}
\begin{aligned}
E(W)_{L1} &= E(W) + \lVert W \rVert_1 \\
E(W)_{L2} &= E(W) + \lVert W \rVert_2 \\
E(W)_{L1+L2} &= E(W) + \lVert W \rVert_1 + \lVert W \rVert_2\\
\end{aligned}
\end{equation}
   
\subsection{Implicit and explicit regularization}
Most widely used and effective explicit regularizers are Data Augmentation, Data Balancing, dropout, l1 regularizer, l2 regularizer.

\emph{Data Augmentation:} Due to increasing in capacity fo the deep network and scarcity of enough discriminant labeled data, it's useful to generate deformed version fo original training examples using affine transformation such as rotation, translation, shearing, mirroring and random cropping. Apart from that color space transformation such as RGB to Lab or HSV, also cropping and resizing proved to be effective in reducing overfitting. 

\emph{Data Balancing:} For a dataset with biased sample classes, learning tends to overfit the class with high bias, results in poor generalization performance. Sample balancing methods such as uniform sampling and stratified sampling resample from data for balanced mini batches as per the given probability distribution of the classes. 

\emph{Dropout:} Dropping layer activation randomly realizes ensemble of many functions for that layer, it helps reducing overfitting.
 
\emph{L1 \& L2 regularizer:} L1 regularizer encourages sparsity by minimizing the L1 norm of the model weights. L2 regularizer penalizes model complexity, leads to small weights. The added combination of L1 \& L2 regularizations encourages sparsity with small weights. The effectiveness of each regularization depends on the application; in general L2 regularization works perfectly fine.

\emph{Normalization:} Batch Normalization (BN) \cite{bn} is one of the de facto implicit regularization for faster and better generalization learning of feedforward deep convolutional model. BN normalizes the layer inputs to a zero mean and unit variance distribution. A model with BN can put an end to the bias term necessity. In the case of recurrent neural network layer normalization \cite{ln} proved to be efficient than batch normalization. Another useful implicit regularizer is Local Response Normalization (LRN). LRN \cite{alex} is inspired from the lateral inhibition of an excited neuron. Unbounded activations are normalized using the values of the local window. For applications such as person re-identification \cite{reid} LRN outperforms BN. 

An experimental analysis of the effectiveness of the batch norm and dropout can be observed from Table \ref{tab:ac_regularization}

\subsection{Feature Space Analysis}
\textbf{Claim 1.3}: The rate of reduction of feature space size with respect to the number of convolutional layers plays important role in generalization.\\
A detailed analysis is given in Section \ref{fp_design} on the impact of feature space design for features representation without losing significance information. Efficient feature space design improves generalization performance by a fair margin.  Table~\ref{tab:acdesign} proves our claim.

\subsection{Learning Rate Decay}
\textbf{Claim 1.4}: Learning rate decay policy acts as an implicit regularizer for deep model learning.\\
Learning rate decay policy plays a major role in generalized parameter learning with faster convergence. As the learning progress, exploration in local neighborhood becomes more important to do away with oscillation and ill-conditioning.
 The Taylor series approxmation of the cost function $f(x)$:
 
 \begin{equation}
 \label{eq:illc}
 \begin{aligned}
 f(x)  &\approx f(x_0) + (x - x_0)^Tg + \frac{1}{2}(x - x_0)^TH(x - x_0) \\
 f(x - \epsilon g) & \approx f(x_0) - \epsilon g^Tg + \frac{1}{2}\epsilon^2g^THg
\end{aligned} 
 \end{equation}
 where $g$ and $H$ are the gradient and the hessian of the cost function. When the values of $H$ are large cost increases, this effect is known as ill conditioning and a common probelm with deep learning training. The learning rate $\epsilon$ decay alleviate the effect of ill conditioning leading to low cost space exploration.
 In practice, polynomial decay works very well in comparison to step decay or inverse decay methods. Step decay needs more supervision, better not to use. 
 
 From Table \ref{tab:ac_policy} we can see that polynomial decay performs much better than step decay. 
 
 \subsection{Optimal Number of Parameters}
 In the deep model impact of VC dimension is marginal. The fat shattering dimension plays important role in regularization. In determining the optimal numbers of parameters feature space design comes into play. Inefficient shallow model or extra deep model may underfit/overfit the data resulting in poor generalization performance. As per the discussion in Section \ref{fp_design}, feature space design plays a major role in powerfull representation on uncorellated hyperspaces reducing information loss. 
 
\textbf{Claim 1.5}: The optimal numbers of parameters, is the number of parameters of a optimal model designed using feature space analysis. \\
Table \ref{tab:ac_depth} and \ref{tab:acdesign} gives an experimental validation of this claim.

\section{Experiment Setup}
\textbf{Platform Details:} All our experiments were carried out on a Linux server with 128GB RAM, Xeon E5-4667 v4 processor, and two Nvidia K80 GPUs.

\begin{table}[!tbp]
\caption{Results of 3 main Designs}
\label{tab:acdesign}
\begin{center}
\begin{tabular}{|c|c|c|}
\hline
Model & \#params(K) & test\_accuracy (\%)\\
\hline
design 1 & 20173  & 89.4 \\
\hline
design 2 & 20173 & 86.8 \\
\hline
design 3 & 20025 & 87.9 \\
\hline
\end{tabular}
\end{center}
\end{table}

 \begin{table}[!tbp]
\caption{Convolution vs Max Pooling}
\label{tab:ac_max_conv}
\begin{center}
\begin{tabular}{|c|c|c|}
\hline
Model & \#params (K) & test\_accuracy (\%)\\
\hline
design 1\_conv & 20948  & 91.7 \\
\hline
design 1 (max\_pooling) & 20173 & 89.4 \\
\hline
\end{tabular}
\end{center}
\end{table}

\begin{table}[!tbp]
\caption{Results of explicit regularization}
\label{tab:ac_regularization}
\begin{center}
\begin{tabular}{|c|c|c|c|}
\hline
Model & dropout & batch\_norm & test\_accuracy (\%)\\
\hline
design 1\_conv & yes & yes & 91.7 \\
\hline
design 1\_conv & yes & no & 88.2 \\
\hline
design 1\_conv & no & yes & 90.1 \\
\hline
\end{tabular}
\end{center}
\end{table}

\begin{table}[!tbp]
\caption{Results of Learning rate decay policy}
\label{tab:ac_policy}
\begin{center}
\begin{tabular}{|c|c|c|}
\hline
Model & policy & test\_accuracy (\%)\\
\hline
design 1\_conv & polynomial  & 91.7 \\
\hline
design 1\_conv & step & 90.1 \\
\hline
\end{tabular}
\end{center}
\end{table}

\begin{table}[!tbp]
\caption{Results of rate of reduction}
\label{tab:ac_reduction}
\begin{center}
\begin{tabular}{|c|c|c|}
\hline
Model & first\_layer\_stride & test\_accuracy (\%)\\
\hline
design 1\_conv & no & 91.7 \\
\hline
design 1\_conv\_stride & yes & 89.4 \\
\hline
\end{tabular}
\end{center}
\end{table}

\begin{table}[!tbp]
\caption{Depth}
\label{tab:ac_depth}
\begin{center}
\begin{tabular}{|c|c|c|}
\hline
Model & \#params (K) & test\_accuracy (\%)\\
\hline
design 1\_conv & 20948 & 91.7 \\
\hline
design 4 & 21573 & 89.3 \\
\hline
\end{tabular}
\end{center}
\end{table}

\begin{table*}[!tbp]
\caption{Network design}
\label{tab:net_design}
\begin{center}
\begin{tabular}{|c|c|c|c|c|c|}
\hline
 block & design 1 & design 1\_conv& design 2 & design 3 & design 4\\
\hline
input & 28 x 28 x 3 & 28 x 28 x 3 & 28 x 28 x 3 & 28 x 28 x 3 & 28 x 28 x 3\\
\hline
block1 & 1 x conv 3x3, 1, 64 & 1 x conv3x3, 1, 64 & 1 x conv3x3, 1, 64 & conv3x3, 64 & 2 x conv3x3, 1, 64\\
\hline
block2 & max\_pool & 1 x conv3x3, 2, 64 & max\_pool & max\_pool & 1 x conv3x3, 2, 64\\
\hline
block2\_1 & - & - & - & - & 1 x conv1x1, 2, 128\\
\hline
block3 & 2 x conv 3x3, 1, 128 & 2 x conv3x3, 1, 128 & 1 x conv3x3, 1, 128 & 1 x conv3x3, 1, 128 & 3 x conv3x3, 1, 128\\
\hline
block3\_1 & - & - & max\_pool & - & block2\_1 + block3\\
\hline
block3\_2 & - & - & 1 x conv3x3, 1, 128 & - & 3 x conv3x3, 1, 128\\
\hline
block4 & max\_pool & 1 x conv3x3, 2, 128 & max\_pool & max\_pool & 1 x conv3x3, 2, 128 \\
\hline
block5 & 4 x conv 3x3, 1, 256 & 4 x conv3x3, 1, 256 & 4 x conv3x3, 1, 256 & 4 x conv3x3, 1, 256 & 4 x conv3x3, 1, 256\\
\hline
block6 & max\_pool & 1 x conv3x3, 2, 256 & max\_pool & max\_pool & 1 x conv3x3, 2, 256 \\
\hline
block7 & 1 x conv 1x1, 1, 4096 & 1 x conv1x1, 1, 4096 &1 x conv1x1, 1, 4096 & 1 x conv1x1, 1, 4096 & 1 x conv1x1, 1, 4096\\
\hline
block7\_1 & dropout & dropout & dropout & dropout & dropout \\
\hline
block8 & 1 x conv 1x1, 1, 4096 & 1 x conv1x1, 1, 4096 & 1 x conv1x1, 1, 4096 & 1 x conv1x1, 1, 4096 & 1 x conv1x1, 1, 4096\\
\hline
block8\_1 & dropout & dropout & dropout & dropout & dropout \\
\hline
block9 & 1 x conv 1x1, 1, 10 & 1 x conv1x1, 1, 10 & 1 x conv1x1, 1, 10 & 1 x conv1x1, 1, 10 & 1 x conv1x1, 1, 10\\
\hline
\end{tabular}
\end{center}
\end{table*}

\textbf{Dataset:} To validate our claims we have used image classification CIFAR10 \cite{cifar10} dataset. It has 10 object classes and divided into two splits for training and validation. The training split has 50000 and the validation split has 10000 images. The size of each image is $32 \times 32 \times 3$, RGB color channels.

\textbf{Preprocessing:} For training a randomly cropped patch of size $28 \times 28 \times 3$ is used. Each patch is flipped left/right and up/down based on coin flipping results. Apart from that, we adjust the image color by scaling its values into $[0, 1]$ range and changing its hue, contrast, and saturation. Each image (training/validation) is standardized by subtracting its mean and dividing its standard deviation. \\
For evaluation, the central crop of each image is selected and resized using bilinear interpolation. 

\textbf{Framework:} We have used TEFLA \cite{tefla}, a python framework developed on the top of TENSORFLOW \cite{tf}, for all experiments described in this work. 

\textbf{Model:} Table~\ref{tab:net_design} details model design for different experiments. Conventions are followed as  ($ repeat \hskip 0.5em \times \hskip 0.5em conv3\times3, \hskip 0.5em, stride, \hskip 0.5em num \_ kernels$); where $repeat$ is the number of convolution for composition design, $stride$ is the stride for convolution and $num\_kernels$ is the number of kernel for each convolution layer. Each convolutional layer of a model is followed by a batch normalization and a non-linearity (relu for ur experiments) layer for all designs experimented in this work.

\subsection{Results Analysis}
Table~\ref{tab:acdesign} shows performance of each model on CIFAR10 dataset validation/test set. For design 1 and design 2 with the same number of parameters, generalization performance varies significantly, asserting the importance of feature space size importance and minimization of information loss.

Table~\ref{tab:ac_max_conv} shows the importance of strided convolution for feature space size reduction than max pooling. Significance performance gain is observed while using convolution instead of max pooling; implying the information loss for max pooling is higher than strided convolution. For design 1\_conv if we use strided convolution for the first layer instead of the second there's a significant drop of generalization performance even though number of parameters remain same, Table~\ref{tab:ac_reduction}. 

Table~\ref{tab:ac_regularization} shows the importance of dropout and batch normalization for generalization. Effect of batch normalization is higher than the dropout.

Table~\ref{tab:ac_policy} proves our claim that learning rate decay policy also acts as implicit regularizer improving generalization performance. polynomial decay is very robust and requires minimal supervision, yielding better generalization performance.  

Performance doesn't always depend on more depth, an optimal design performs better than a deeper design, from Table~\ref{tab:ac_depth} we can see the experimental results of two design. From this, we can conclude that there exists an optimal number of parameters for generalization.

\section{Conclusion}
In this work, a detailed analysis of deep model generalization performance trade-off is presented. We showed that the compositional feature space design with implicit and explicit regularizations play important role in achieving better performance. In terms of model complexity traditional measure, VC dimension doesn't give much information, but fat-shattering dimension analysis has an indirect effect on generalization. From our experiment, we showed that the optimal model satisfies compositional design criteria and have the optimal number of parameters. We wrap up this work with the claim that combination of compositional feature space design with explicit and implicit generalization and efficient optimization algorithms give the best-generalized performance for any dataset.






%

\section{Appendix}

\subsection{Effective receptive field}
\begin{equation}
\begin{aligned}
j_{out} = j_{in} * s \\
r_{out} = r_{in} + (k - 1)*j_{in}
\end{aligned}
\end{equation}
where $j$ is the distance between two adjacent feature maps; $s$ is the convolution stride, $k$ is the kernel size and $r$ is the receptive fields size.

\subsection{Learning rate decay policy}
Commonly used learning rate decay policies are given below:

\begin{align}
\begin{split}
fixed: & \lambda = c 
\end{split} \\
\begin{split}
exponential: &\lambda_{iter} = \lambda_0 * \gamma^{iter}
\end{split} \\
\begin{split}
step: &  \lambda_{iter} = \lambda_0 * \gamma^{\frac{iter}{step}}
\end{split} \\
\begin{split}
inverse: & \lambda_{iter} = \lambda_0 * (1 + \gamma * iter)^{-c}
\end{split} \\
\begin{split}
poly: & \lambda_{iter} = \lambda_0 * (1 - \frac{iter}{max\_iter})^{c}
\end{split} \\
sigmoid: & \lambda_{iter} = \lambda_0 * \frac{1}{1 + exp^{(-\gamma + (iter - step))}}
\end{align}

where $c$ and $\gamma$ are two constants; $\lambda_0$ is the initial learning rate, $\lambda_{iter}$ is the learning for $iter$ (current iteration). $max\_iter$ is the maximum number of iteration for learning and $step$ is the step for changing learning rate for step policy. 

\subsection{Mutual information for normal random variables}
For a normal random variables X $\sim N(\mu, C)$, entropy of X is calculated as follows

\begin{flalign*}
H(X) &= -Xlog_{a}(X) &&\\
&= - \int p(x)log_{a}p(x) dx &&\\
&= \int p(x)[\frac{1}{2}log_{a}(2\pi)^n\lvert C \rvert + \frac{1}{2}(x - \mu)^TC^{-1}(x-\mu)log_{a}e]dx&&\\
&= \frac{1}{2}log_{a}(2\pi)^n\lvert C \rvert + \frac{1}{2}log_{a}e E[(x - \mu)^TC^{-1}(x-\mu)] &&\\
&= \frac{1}{2}log_{a}(2\pi)^n\lvert C \rvert + \frac{1}{2}nlog_{a}e&&\\
&= \frac{1}{2} log_{a}(2\pi e)^{n}\lvert C \rvert &&
\end{flalign*}
 For another normal random variables Y $\sim (N\mu, C_{Y})$; the mutual information between X and Y can be calculated as follows:
 \begin{flalign*}
 I(X; Y) &= H(X) - H(X|Y)&&\\
 &=H(X) + H(Y) - H(X, Y) &&\\
 &= \frac{1}{2} log_{a}(2\pi e)^{n}\lvert C \rvert + \frac{1}{2} log_{a}(2\pi e)^{n}\lvert C_{Y} \rvert - \frac{1}{2} log_{a}(2\pi e)^{n}\lvert C_{XY} \rvert &&\\
 &= \frac{1}{2} log_{a}(2\pi e)^{n}\frac{\lvert C \rvert \lvert C_{Y} \rvert}{\lvert C_{XY} \rvert} && \\
 \end{flalign*}

\end{document}